\documentclass[10pt, a4paper]{article}

\usepackage[final]{lrec2026} 
\usepackage{amsmath}
\usepackage{svg}
\usepackage{multirow}
\usepackage{booktabs,tabularx}
\usepackage{subfig}

\title{From Variance to Invariance: Qualitative Content Analysis for Narrative Graph Annotation}

\name{Junbo Huang$^{\ast}$, Max Weinig$^{\dagger}$, Ulrich Fritsche$^{\dagger}$, Ricardo Usbeck$^{\ddagger}$} 

\address{$^{\ast}$Department of Computer Science, University of Hamburg  \\ junbo.huang@uni-hamburg.de \\ 
          $^{\dagger}$Department of Socioeconomics, University of Hamburg \\  \{max.weinig, ulrich.fritsche\}@uni-hamburg.de \\ 
          $^{\ddagger}$AI and Explanability Group, Leuphana University of Lüneburg \\ ricardo.usbeck@leuphana.de}

\abstract{
Narratives in news discourse play a critical role in shaping public understanding of economic events, 
such as inflation. Annotating and evaluating these narratives in a structured manner remains a key 
challenge for Natural Language Processing (NLP). In this work, we introduce a narrative graph annotation 
framework that integrates principles from qualitative content analysis (QCA) to prioritize annotation 
quality by reducing annotation errors. 
We present a dataset of inflation narratives annotated as directed acyclic graphs (DAGs), 
where nodes represent events and edges encode causal relations. 
To evaluate annotation quality, we employed a $6\times3$ factorial experimental design to examine the 
effects of narrative representation (six levels) and distance metric type (three levels) on inter-annotator 
agreement (Krippendorrf's $\alpha$), capturing the presence of human label variation (HLV) in narrative 
interpretations. Our analysis shows that (1) lenient metrics (overlap-based distance) overestimate reliability, and
(2) locally-constrained representations (e.g., one-hop neighbors) reduce annotation variability. 
Our annotation and implementation of graph-based Krippendorrf's $\alpha$ are open-sourced. 
The annotation framework and evaluation results provide practical guidance for NLP research on 
graph-based narrative annotation under HLV. 
 \\ \newline  }

\begin{document}

\maketitleabstract

\section{Introduction}

In recent years, there has been growing interest in the study of narratives in economics~\citep{shillerNarrativeEconomics2017, roosNarrativesEconomics2023} and social sciences~\citep{pollettaSociologyStorytelling2011, beckertUncertainFuturesImaginaries2018}. Narratives are of central importance in these fields because they ``[...] give meaning to life’s activities, to sense truth, and to create the commitment to act''~\citep[xvii]{tuckettMindingMarkets2011}, and they may trigger shifts or turning points that drive economic and societal dynamics~\citep{roosNarrativesEconomics2023}. Within the literature they are commonly defined as a ``sequence of causally linked events''~\citep{akerlofAnimalSpiritsHow2010} and are formally represented as Directed Acyclic Graphs (DAGs)~\citep{eliazModelCompetingNarratives2020}.

As an important application of Natural Language Processing (NLP), identifying and extracting narratives from large text corpora is a central objective in many interdisciplinary research areas. In this paper, we present an inflation narrative dataset constructed from news articles,  with a strong focus on ensuring a transparent and rigorous approach to \textbf{narrative graph annotation and evaluation}.

To begin with, we identify three major challenges in narrative annotation and its evaluation from an interdisciplinary perspective. First, \textbf{annotating narratives presents additional challenges compared to standard sequence classification or labeling tasks in NLP}. Narrative annotation involves identifying and linking events across larger text segments. This requires interpretive judgments that are often subjective and context-dependent, making consistency and reproducibility more difficult to achieve with conventional annotation frameworks. Integrating qualitative methodologies such as Qualitative Content Analysis (QCA), widely used in the humanities and social sciences, can address this challenge by providing a systematic procedure for interpreting and validating complex narrative structures. Yet their adoption in NLP remains limited. 

\begin{figure}
\centering
\includegraphics[width=0.5\textwidth]{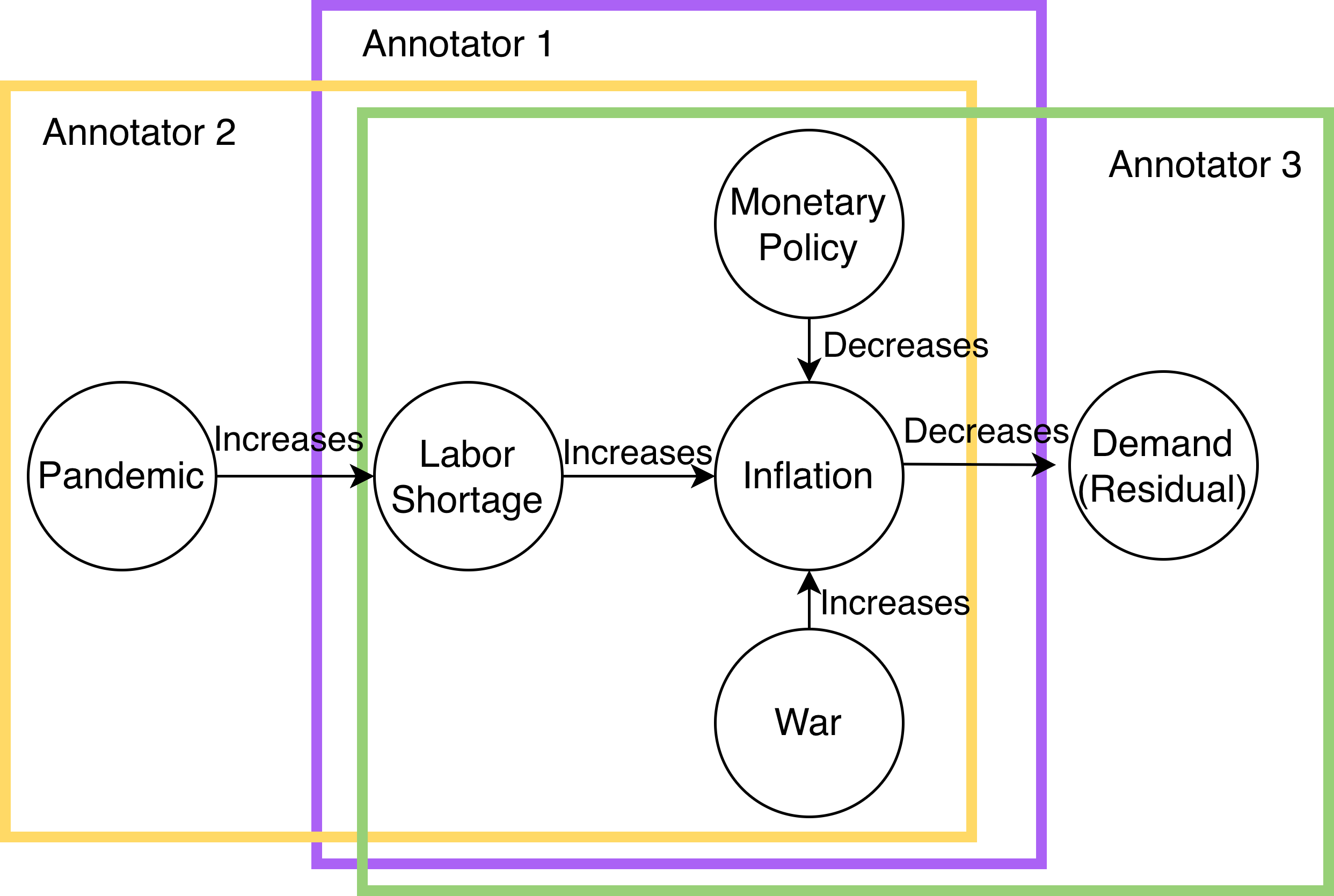}
\caption{Example of narrative graph annotation under human label variation.}
\label{fig1}
\end{figure}

Second, as in Fig.~\ref{fig1}, \textbf{graph-based representations introduce substantial variation across annotators}. Unlike flat representations of narratives, such as assigning a single label to a sequence, graph-based structures provide greater degrees of freedom in how events are connected. Annotators may differ in identifying relevant events, choosing relation types, or determining graph granularity, all of which can lead to divergent yet plausible annotations. Research on human label variation (HLV) indicates that ``there exists genuine human variation in labeling due to disagreement, subjectivity in annotation or multiple plausible answers''~\citep{DBLP:conf/emnlp/Plank22}.

Third, \textbf{there is no established consensus on how to assess inter-annotator agreement (IAA) for narrative graph annotations}. Existing graph distance metrics, such as graph edit distance, capture different structural or semantic dimensions of similarity, and their relevance often depends on the specific analytic goals or application domains. This lack of standardization complicates the interpretation of agreement scores, particularly in the presence of HLV. To address this, we compute multiple IAA measures based on diverse distance metrics and narrative representations, following the comparative evaluation approach proposed by ~\citet{DBLP:conf/lrec/Passonneau06}.

To our knowledge, no previous work has jointly addressed these challenges in narrative annotation and evaluation. \textbf{Our contributions are threefold}: (1) we propose a QCA-based methodology for narrative graph annotation; (2) we develop an graph-based annotation evaluation framework that accounts for HLV, and (3) we identify a reliable graph representation that balances contextual completeness with annotation consistency, capturing core narrative elements while maintaining robust reliability. We ask the following questions.

\begin{itemize}
    \item[\textbf{RQ1:}] How can QCA enhance the reliability and methodological rigor of narrative graph annotation with the presence of HLV?
    \item[\textbf{RQ2:}] How does the choice of narrative representation (categorical vs. graph-based) and distance metric (lenient vs. moderate vs. strict) affect reliability score?
    \item[\textbf{RQ3:}] Which graph representation best captures reliable and commonly-perceived narrative elements while maintaining contextual completeness? 
\end{itemize}

\section{Preliminaries}
We briefly introduce narrative representation, QCA, and the computation of Krippendorff’s alpha for IAA.

\subsection{Inflation Narratives as Graphs}

We conceptualize news discourse narratives as structured representations of events and their causal linkages. Conversely, early work in economics and social sciences often relied on broad definitions, treating narratives as ``stories that offer interpretations of economic events, morals, or hints of theories about the economy''~\citep{shillerNarrativeEconomics2017}, and operationalized them through topics or sentiment-based features~\citep{müllerGermanInflationNarrative2022, terellenNarrativeMonetaryPolicy2022, Weinig2025Going}. While these approaches capture general trends, they offer limited insight into the structural complexity of narratives.

More recent methods aim to structurally identify core narrative elements (e.g., agents, relations, actions, outcomes) to better represent narratives~\citep{gehringAnalyzingClimateChange2023, ashRelatioTextSemantics2024, gueta-etal-2025-llms}. However, most still overlook interpretive variation, offering limited insight into the subjective nature of narrative interpretation, where multiple plausible readings of the same text coexist.

Furthermore, these approaches lack alignment with the formal definition of narratives grounded in the DAG framework~\citep{Pearl.2009}, as introduced by ~\citet{eliazModelCompetingNarratives2020}. Following this idea, they define narratives, in line with the literature studies definitions of narratives~\citep{abbottCambridgeIntroductionNarrative2002}, as a series of events that are causally linked. This formalization enables narratives to be represented as graphs, where \textbf{nodes correspond to events and edges encode directed causal relations}~\footnote{We use the term node instead of vertex throughout the paper for simplicity}.

To this end, we grounded our narrative representation as DAGs. In the context of text-to-graph annotation, the annotated DAGs, in contrast to ~\citet{heddaya-etal-2024-causal}, are not restricted in size. Given the presence of HLV in narrative annotation, \textbf{we address having multiple plausible annotations not as an issue, but a signal to reveal the commonly-perceived story within a news article}.

\subsection{Qualitative Content Analysis}

\begin{figure}
\centering
\includegraphics[width=0.35\textwidth]{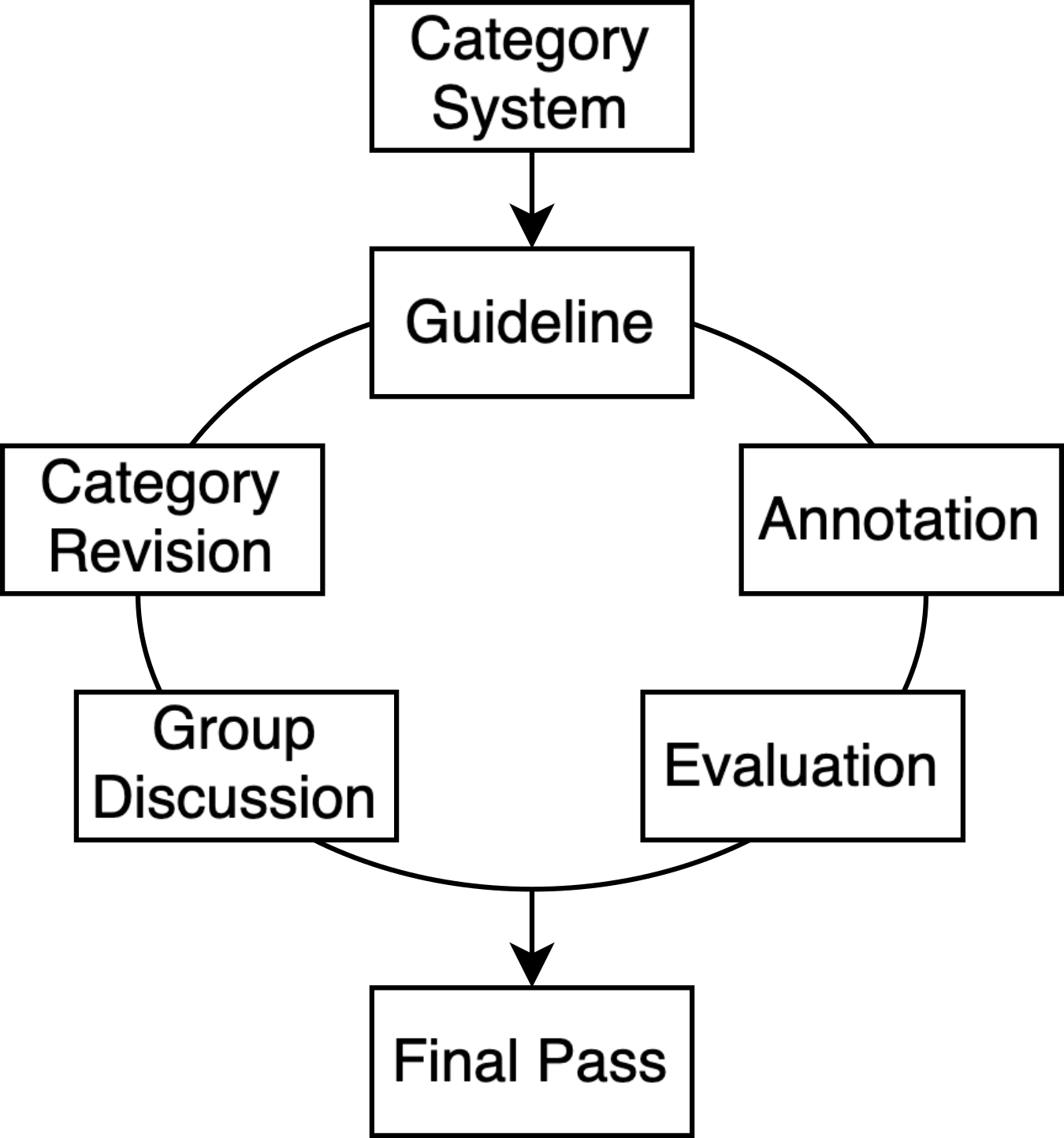}
\caption{Flowchart of iterative QCA research procedure}
\label{fig2}
\end{figure}

Qualitative methods commonly used in social sciences and humanities have the ability to capture rich and complex information, including manifest content and latent meaning. One widely used empirical method for text analysis is QCA~\citep{selviQualitativeContentAnalysis2019}. Unlike other interpretative methods, QCA follows a rule-based procedure that ensures a high degree of inter-subjective verifiability, making it a natural candidate for narrative annotation under HLV. Unlike formal annotation frameworks in NLP, QCA is a context-sensitive method that derives coding categories from both theoretical considerations and the textual material itself. In Fig.~\ref{fig2}, we provide a flow diagram that explains our procedure. The central unit of analysis is the category system, which is deductively derived from existing theoretical and empirical research~\citep{Mayring2014}. 
Our category system is primarily based on the work of ~\citet{10.1093/restud/rdag014}, who study existing backward-looking inflation narratives among households and experts in the US. They group these narratives into three supercategories (supply, demand and miscellaneous) with their corresponding subcategories.

Following ~\citet{przyborksiForschungsdesignFuerQualitative2022}'s understanding of qualitative research as an open and circular process, we inductively extend and adjust the initial category system during the annotation process (Fig.~\ref{fig2}). Acknowledging temporal and contextual differences in narrative structures, we iteratively develop additional narrative subcategories through regular group discussions involving annotators and the research team. Additionally, we refine our coding guidelines throughout the annotation process to reduce annotation errors and improve consistency. Finally, we arrived at 26 fine-granular subcategories (five for \textbf{demand}\footnote{Demand: Government Spending, Monetary Policy, Pent-up Demand, Demand Shift, Demand (Residuals).}, 9 for \textbf{supply}\footnote{
Supply: Supply Chain Issues, Transportation Costs, Labor Shortage, Wages, Energy Prices, Food Prices, Housing Costs, Supply (Residual).}, and 12 for \textbf{miscellaneous}\footnote{
Miscellaneous: Pandemic, Politics, War, Inflation Expectations, Base Effect, Government Debt, Tax Increases, Exchange Rates, Medical Costs, Education Costs, Climate Crisis, Price-Gouging.}). The final category system, including explanation of each subcategory is provided in the appendix.

To sum up, QCA is a systematic, human-centered approach to text interpretation that integrates categorical coding with iterative refinement, supporting transparent and theoretically-grounded annotation of complex, context-dependent phenomena such as narratives. \textbf{In our work, we applied QCA during a pilot phase to refine both the category system and annotation guidelines. In the final annotation stage, QCA-based group discussions facilitated a shared understanding of the iteratively-refined category system among annotators.}

\subsection{Krippendorff's Alpha for Graphs}

To measure IAA rigorously, we opt for the Krippendorff’s alpha ($\alpha$). It is a reliability coefficient which accounts for chance agreement and can handle missing data~\citep{krippendorff2011computing}. In contrast to Cohen's kappa, it can handle any number of annotators and does not assume equal distribution across labels~\citep{parfenova-etal-2025-text}.

The general form for $\alpha$ is:
\begin{equation}
    \alpha=1-\frac{D_o}{D_e}
\end{equation}
, where $D_o$ is the observed disagreement among annotators per annotation item (i.e, per document) and $D_e$ denotes the expected total disagreement within- and between-item. To capture disagreement denoted by $D_o$ and $D_e$, well-defined graph distance metrics are required. For the detailed computation of $D_o$ and $D_e$, we refer readers to ~\citet{krippendorff2011computing}. 

The value of $\alpha$ ranges from -1 to 1, where $\alpha < 0$ implies a systematic disagreement, likely caused by annotators not instructed well for the annotation task. $\alpha=0$ suggests no agreement beyond chance, and $\alpha>0$ the observed agreement is beyond chance-level (e.g., annotators understand the annotation task well and have some degree of agreement). It is worth mentioning that values assigned to items can be categorical, ordinal, interval, or ratio, and any distance metric can be applied. However, to our best knowledge, no publicly available implementation of Krippendorff's $\alpha$ exists for graph annotations. To this end, we modified the current implementation of $\alpha$ and extended its computation for graph features, which includes distance metrics for nodes, edges and graphs\footnote{We open-source the implementation of Krippendorff's $\alpha$ for graph under \url{https://pypi.org/project/krippendorff-graph/}}. Formally, we represent narrative graphs as a set of subject-predicate-object triples, e.g., $\{\{sub_1, pred_1, obj_1\}, \{sub_2, pred_2, obj_2\}\}$. The following distance metrics are considered.

\subsubsection{Lenient Distance Metric}
\label{sec:lenient}
\paragraph{Node/Edge Overlap Distance}
As shown in Eq.~\ref{eq2}, given two sets of nodes or edges (a set of strings), $n_1$ and $n_2$, node/edge overlap distance measures if these two sets overlap. The measurement of set overlap is a lenient measurement of distance.

\begin{equation}\label{eq2}
    d_{lenient}(n_1, n_2)= 
\begin{cases}
    0,              & \text{if } |n_1\cap n_2| > 0 \\
    1,              & \text{otherwise}
\end{cases}
\end{equation}

where $|\cdot|$ denotes the cardinality of a set.

\paragraph{Graph Overlap Distance}
Similar to node/edge overlap distance, graph overlap distance (Eq.~\ref{eq4}) measures if two graphs (a set of triples), $g_1$ and $g_2$, overlap. 

\begin{equation}\label{eq4}
    d_{lenient}(g_1, g_2)= 
\begin{cases}
    0,& \text{if } |g_1 \cap g_2|>0 \\
    1,              & \text{otherwise}
\end{cases}
\end{equation}

\subsubsection{Moderate Distance Metric}
\label{sec:moderate}
\paragraph{Node/Edge Jaccard Distance}
To capture partial similarity, we adopt the Jaccard distance, which measures the proportion of shared nodes or edges between two sets (Eq.~\ref{eq:jaccard}). It provides a graded measure of agreement, reducing overestimation from simple overlap.

\begin{equation}\label{eq:jaccard}
d_{moderate}(n_1, n_2) = 1 - \frac{|n_1 \cap n_2|}{|n_1 \cup n_2|}
\end{equation}

The Jaccard distance ranges from 0 (identical) to 1 (no overlap).

\paragraph{Graph Edit Distance}
Eq.~\ref{eq5} defines this standard similarity measure between two graphs.

\begin{equation}\label{eq5}
    d_{strict}(g_1, g_2) = \min_{(e_1,\dots e_k)\in\mathcal{P}(g_1, g_2)}\sum_{i=1}c(e_i)
\end{equation}

$\mathcal{P}(g_1, g_2)$ represents the set of edit operations required to transform graph $g_1$ into $g_2$. Each edit operation $e_i$ can be a node or edge insertion, deletion, or substitution. The cost function $c(e_i)$ assigns a constant cost of 1 to each edit, and the total distance corresponds to the minimal number of edits needed to make the two graphs identical.

Because annotated graphs can vary substantially in size, normalization is necessary to ensure scale invariance. The raw edit distance between $g_1$ and $g_2$ is therefore normalized by the sum of their respective distances to an empty graph ($g_0$), yielding a bounded score between 0 and 1 (Equation~\ref{eq6}).

\begin{equation}\label{eq6}
    \hat{d}_{strict}(g_1, g_2) = \frac{d_{strict}(g_1, g_2)}{d_{strict}(g_1, g_0)+d_{strict}(g_2, g_0)}
\end{equation}

\subsubsection{Strict Distance Metric}
\label{sec:strict}
\paragraph{Node/Edge Exact Match}
Eq.~\ref{eq3} defines the exact match between two sets of nodes or edges. $d_{strict}(n_1, n_2)$ assigns a distance of 0 only when the two sets are identical; if any element differs between them, the distance is 1.

\begin{equation}\label{eq3}
    d_{strict}(n_1, n_2)= 
\begin{cases}
    0,& \text{if } n_1 = n_2 \\
    1,              & \text{otherwise}
\end{cases}
\end{equation}

\paragraph{Graph Exact Match}
For the strictest comparison, we compute graph-level exact match, which requires two graphs to be structurally identical—that is, they must contain the same set of triples (subject, relation, object) with identical labels. As shown in Equation~\ref{eq:graphem}, any missing, extra, or mismatched node or edge results in a maximal distance of 1.

\begin{equation}\label{eq:graphem}
d_{strict}(g_1, g_2)=
\begin{cases}
0,& \text{if } g_1 = g_2 \\
1, & \text{otherwise}
\end{cases}
\end{equation}

This metric captures complete structural equivalence between two narrative graphs and penalizes even minor discrepancies, providing a conservative lower bound on inter-annotator reliability. While it highlights exact agreement, it may underestimate consistency in cases where annotators capture the same underlying causal structure using slightly different formulations or graph sizes.

\section{Narrative Graph Annotation}
We detail the dataset sampling and filtering criteria, and outline the narrative annotation task as well as 
the annotator recruitment process. Furthermore, we explain how the annotation guidelines and category system 
were iteratively refined through a QCA-based pilot study to minimize annotation errors. Note that this QCA-based 
annotation methodology is best suited when annotation quality is the top priority, especially for annotation tasks 
where human label variation (i.e., disagreement, subjectivity and multiple plausible answers) is of interest. 
Additionally, the discussion around reproducibility should be focused on the annotation methodology 
as well as evaluation methodology, rather than the dataset itself, which is unfortunately not open-sourced 
due to licensing restrictions.

\subsection{Data Source}

We use the English news corpus from the Dow Jones Newswires Machine Text Feed and Archive (DJN)\footnote{Unfortunately DJN is a licensed corpus. Therefore we can not open-source the original text. We nevertheless open-sourced the annotation and analyses: \url{https://github.com/semantic-systems/LREC2026-From-Variance-to-Invariance}.} database. The dataset includes content from the following areas: market-moving M\&A, exclusives, and earnings news; full-text feeds from Dow Jones sources (Newswires, Barron's, MarketWatch); global company news; central bank, macroeconomic, political, FX, commodities, and energy news; and third-party press release wires (BusinessWire, PR Newswire, Globe Newswire, among others). Notably, the corpus includes content from one of the largest US newspapers, \textit{The Wall Street Journal}.

\subsection{Tasks Description}

Our annotation process is two-fold. We begin with a document-level classification task. This is followed by an extraction task, which involves highlighting event spans and relations.

\subsubsection{Task 1: Narrative Identification}

To determine whether a document addresses the cause(s) of inflation, Task~1 is a classification task with the following three classes: 
\begin{enumerate}
    \item Inflation-cause-dominant: The document discusses the cause(s) of inflation.
    \item Inflation-related: The document mentions inflation but not inflation cause(s). 
    \item Non-inflation-related: The document neither mentions causes of inflation nor inflation.
\end{enumerate}

\subsubsection{Task 2: Narrative Extraction}

The second annotation task consists of two steps. First, annotators mark relevant event spans using the derived category system. Second, annotators identify the causal relation between the marked events (i.e., \textit{Increases} or \textit{Decreases}). Although annotation is performed at the span level, the resulting representation is aggregated at the document level, abstracting from the exact span boundaries. 
Appendix~\ref{app:example} provides an example with annotations.

\subsection{Annotator Recruitment}

Given the complexity of the task and the importance of domain-specific knowledge, we recruited students with academic training in economics to carry out the data annotation. In total, seven students participate in the study, three in the pilot phase (one cisgender female and two cisgender males\footnote{All genders are self-identified.}) and four in the final annotation phase (two cisgender females and two cisgender males). All annotators have studied economics, at least during their bachelor’s degrees, and five out of the seven are enrolled in graduate programs with a substantial economics component at the time of the study. Four of them have experience with QCA through their academic coursework. All annotators were between 24 and 30 years old, and two out of seven reported having a migration background. The hourly wage is €14.37 for the pilot study, and €18.30 for the final study.
Further details are provided in Appendix~\ref{app:annotators}.

\subsection{Pilot Study}
\label{pilot_study}
We began with a simple assumption: \textbf{fewer annotation errors should lead to higher IAA}, even accounting for inevitable differences in interpretation. To refine our setup and minimize errors, we conducted a pilot study\footnote{Conducted from August to September 2024.} within a QCA framework.

We identified three main sources of annotation errors:  
(i) \textbf{under-specified guidelines}, which left annotators uncertain about decision boundaries;  
(ii) \textbf{misunderstandings of the category system}, where annotators misinterpreted correct definitions or failed to apply them as intended; and  
(iii) \textbf{ambiguities in class definitions within the category system itself}, where the wording allowed multiple reasonable interpretations even for careful readers.
Rather than explicitly measuring annotation errors or separating them from inherent human variation, we focused on iterative improvement of the annotation setup. After each annotation round, annotators participated in group discussions to resolve misunderstandings, clarified ambiguous definitions, and aligned on task interpretations, guideline usage, and the category system. The pilot study targeted four design questions:
\begin{itemize}
    \item \textbf{Q1 (Sampling strategy):} Does the initial \textit{temporally equal-weighted stratified sampling} approach\footnote{Samples are drawn by dividing the corpus into time-based strata and randomly selecting documents within each stratum.} yield enough articles primarily focused on \textit{causes} of inflation?
    \item \textbf{Q2 (Annotation workflow):} How does performing Task~2 immediately after Task~1 for each document compare to completing Task~1 for all documents before selecting a subset for Task~2?
    \item \textbf{Q3 (Category System):} Which categories in the category system are ambiguous or difficult to apply consistently?
    \item \textbf{Q4 (Span annotation):} How does document length affect annotation quality in identifying relevant spans, and can pre-annotation (e.g., automated highlighting of candidate spans) improve accuracy and reduce annotator fatigue?
\end{itemize}
We have the following findings:
\paragraph{Q1:} Annotators reported that temporally equal-weighted stratified samples produced mostly non-inflation-related documents. 
This bias likely reflects the tendency for inflation-related coverage to surge during price hikes. 
To address this, we revised the strategy to sample only from \textit{inflation-peak years}\footnote{Inflation-peak years include 1990, 1991, 1999, 2001, 2007, 2008, 2021, 2022 and 2023} between 1990 and 2023 (identified using Consumer Price Index data) (Figure~\ref{app:fig1}). 
More descriptive statistics of the sampled documents are provided in the Appendix~\ref{app:descriptives}.
While this increased the volume of inflation-related documents, many were still not cause-dominant. 
We then applied LLaMA-3.1\footnote{\url{https://huggingface.co/meta-llama/Llama-3.1-8B}} for 
zero-shot classification as a filtering step, which raised the average proportion of documents 
annotated with \textit{inflation-cause-dominant} from $\sim$10\% to $\sim$40\% across annotators.

\paragraph{Q2:} Increasing the share of cause-dominant documents did not guarantee sufficient annotation overlap for Task~2 annotations. Under the original \textit{sequential labeling} design, annotators can proceed to Task~2 if they label a document from Task~1 as cause-dominant. With a small annotator pool, this leads to sparse Task~2 coverage: only 31.3\% of 99 pilot documents have three annotations.
To resolve this, we adopted a \textit{batch labeling} procedure: first, all documents are fully annotated for Task~1. Then the most frequently cause-dominant documents are selected for Task~2. 

\paragraph{Q3:} Annotators repeatedly reported ambiguity in distinguishing between the \textit{Pent-up Demand} and \textit{Demand (Residual)} categories, as well as between \textit{Government Spending} and \textit{Government Debt}. These ambiguities informed targeted discussion and revisions to the coding guideline. Moreover, annotators identified cases where a finer-grained category system was required, as many such events frequently appeared in the corpus. Specifically, we introduced additional event categories \textit{Transportation Costs}, \textit{Wages}, \textit{Food Prices}, \textit{Housing Costs}, \textit{Trade Balance}, \textit{Exchange Rates}, \textit{Medical Costs}, \textit{Education Costs}, and \textit{Climate Crisis}, which are novel relative to the original category system by ~\citet{10.1093/restud/rdag014}. We also renamed the labels \textit{Energy Crisis} and \textit{War in Ukraine} to \textit{Energy Prices} and \textit{War}, enhancing temporal generalizability. Finally, we introduced the \textit{Wages} category to capture a broader range of labor market dynamics beyond \textit{Labor Shortage}, while retaining the latter to support a more fine-grained classification.

\paragraph{Q4:} Annotators reported that annotating long documents (some documents are above 1000 words) made it harder to identify precise spans, with a tendency to either overlook relevant segments or select overly broad spans. On average, each document required approximately 15 minutes to annotate in Task~2, which is longer than the initially-estimated 10 minutes, further motivating the need to reduce cognitive load during annotation. To address this, we introduced a pre-annotation step using the Gliner model\footnote{\url{https://huggingface.co/EmergentMethods/gliner_large_news-v2.1}}~\citep{DBLP:conf/naacl/ZaratianaTHC24}. Gliner is a BERT-based, zero-shot, similarity-driven entity extraction model trained on large news corpora. At inference time, it computes distance metrics between entity-type embeddings and mention embeddings, enabling entity extraction for arbitrary sets of entity types without task-specific fine-tuning.
Annotators were free to ignore or modify the automatically highlighted spans, which served purely as navigational aids rather than prescriptive labels. In the final annotated graphs, 41.08\% of nodes were manually added by annotators, reflecting both the model’s bias toward named entities and its imperfect coverage. Annotators were explicitly encouraged to challenge Gliner’s suggestions and to mark additional spans when relevant content was missed. According to annotator feedback, this pre-annotation strategy effectively reduced the cognitive effort while preserving independent interpretive judgment.

\begin{figure*}[!ht]
\centering
\subfloat[All Events]{\includegraphics[scale=0.12]{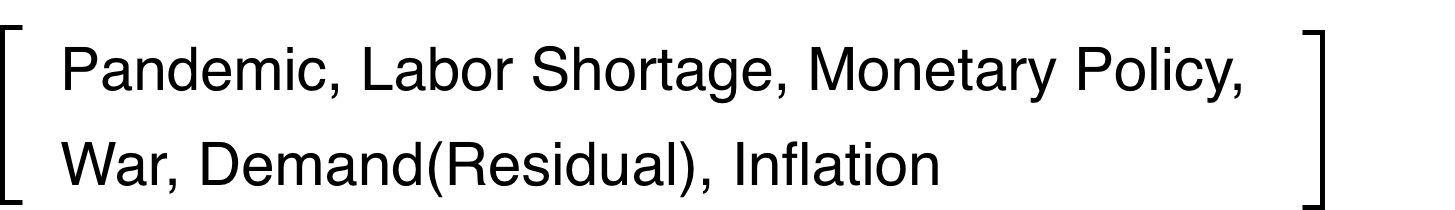}}\qquad
\subfloat[Adjacent Events]{\includegraphics[scale=0.12]{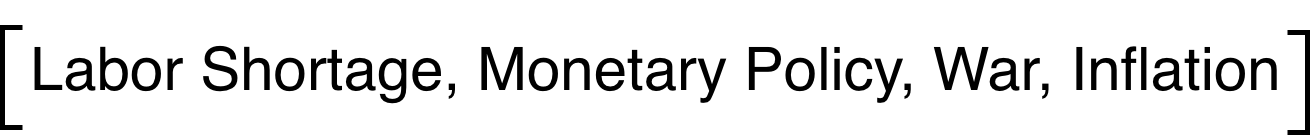}}\qquad
\subfloat[Relations]{\includegraphics[scale=0.12]{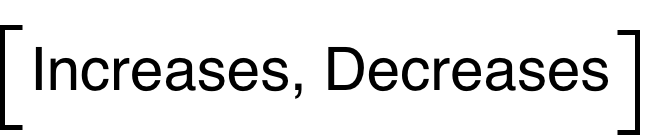}}
\\ 
\subfloat[Full Story]{\includegraphics[scale=0.58]{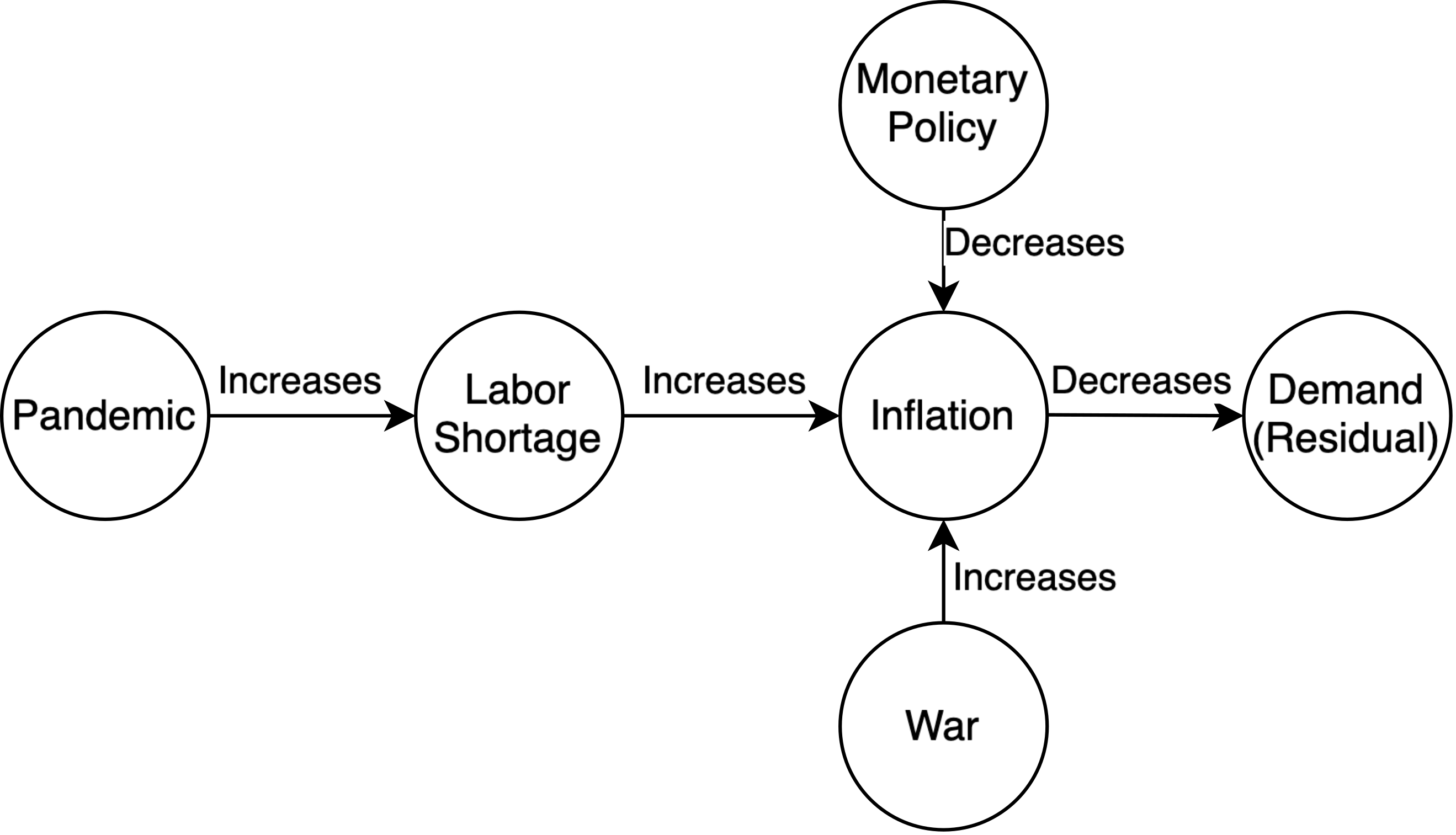}}\qquad
\subfloat[Adjacent Story]{\includegraphics[scale=0.58]{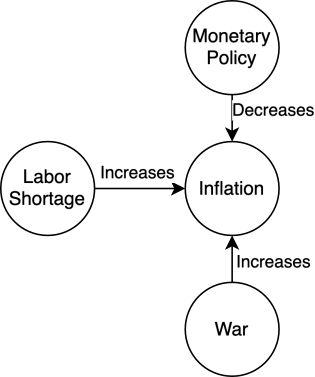}}\qquad
\subfloat[Extended Story]{\includegraphics[scale=0.58]{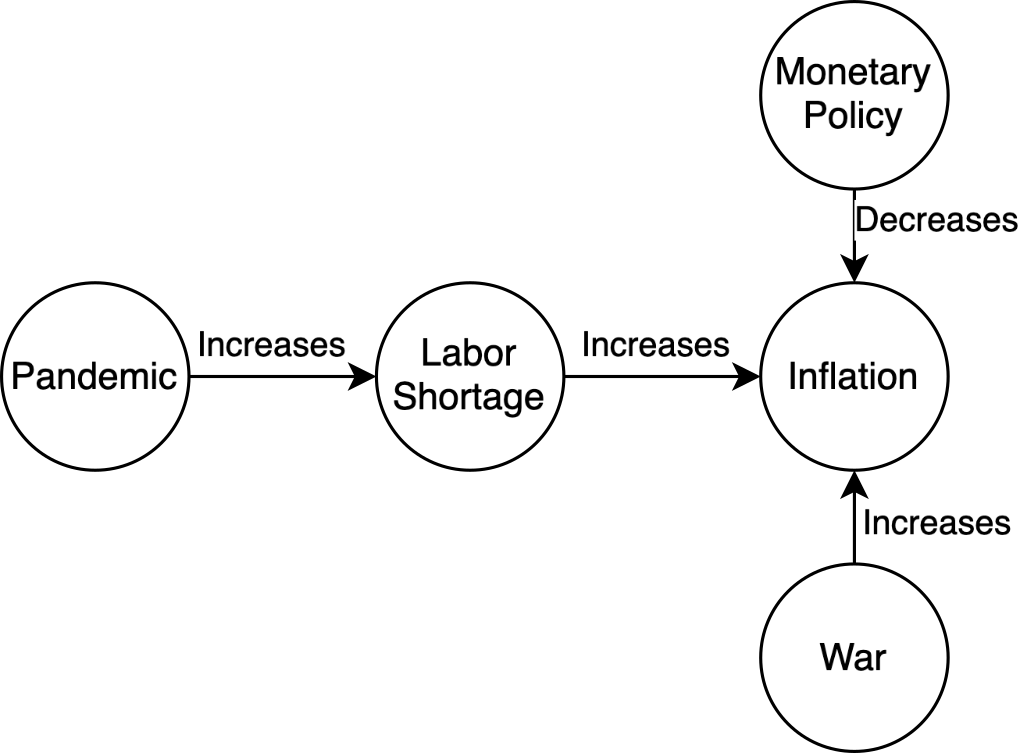}}\qquad 
\caption{Visual example of six types of narrative representations. 
(a-c) represent the categorical representation of narratives, where 
(a) includes all annotated events causing and affected by \textit{Inflation}, 
(b) includes only events directly causing \textit{Inflation} and 
(c) includes a set of relations (\textit{Increases} or \textit{Decreases}) from (e). 
(d-f) represent the graph-based representation of narratives, where 
(d) includes all annotated events and their relations \textit{from} and \textit{to Inflation}, 
(e) includes only annotated events directly causing \textit{Inflation}, as well as their relations, and 
(f) includes only annotated events directly and indirectly causing \textit{Inflation}, as well as their relations.}
\label{fig3}
\end{figure*}

\section{Annotation Evaluation}
For Task~1, we present a standard annotation evaluation practice with a single IAA score (Krippendorff’s $\alpha$). For Task 2, we employed a $6\times3$ factorial experimental design to evaluate the effects of two independent variables on Krippendorff’s $\alpha$: (1) narrative representation and (2) distance metric type. This design yielded 18 distinct reliability scores, enabling a systematic comparison of how both factors influence annotation reliability.

\subsection{Task 1: Narrative Identification}

Turning to Task~1, which involves classification across three labels, we note that no missing or invalid values were recorded by the four annotators. Krippendorff's alpha ($\alpha=0.668$) is computed with a nominal distance metric, where distance between two annotations is 0 if and only if they are identical. The acquired $\alpha$ indicates a moderate agreement.

Instead of developing an optimal model for identifying narratives mentioning causes of inflation, the goal of Task~1 is to select a subset of the annotated sample ($N = 488$) for which annotators consistently agree that inflation causes are explicitly reported by the media.
Among the 488 documents, there are 227 instances where all four annotators assigned the same label from the label set: 33 for \textit{inflation-cause-dominant}, 37 for \textit{inflation-related}, and 157 for \textit{non-inflation-related}. 261 documents depict disagreement. Of these 261 disagreed documents, 202 instances have a unique dominant label via majority voting (71 for \textit{inflation-cause-dominant}, 84 for \textit{inflation-related}, and 47 for \textit{non-inflation-related}). In cases without a clear majority, it is assigned to \textit{inflation-related}. We concatenated these 33 documents labeled as \textit{inflation-cause-dominant} by all four annotators and 71 majority-voted \textit{inflation-cause-dominant} documents. These 104 documents are the basis for Task~2.

\subsection{Task 2: Narrative Extraction}

\begin{table*}[!ht]
\centering
\begin{tabular}{ccccccccccc}
\hline
    &   & \multicolumn{3}{c}{\rule{0pt}{1em}Distance Metric Type} & \multirow{2}{*}{$\Delta_{lm}$}&\multirow{2}{*}{$\Delta_{ms}$}&\multirow{2}{*}{$\Delta_{ls}$}& \multirow{2}{*}{$\mu(|\cdot|)$}& \multirow{2}{*}{$sdt(|\cdot|)$} \\ 
    \multicolumn{2}{c}{Narrative Representation}  & Lenient & Moderate   & Strict     &           \\ \hline
    \multirow{3}{1.5cm}{Category} & \rule{0pt}{1em}All Events &\textbf{0.868}&0.481&0.244& 0.387 & 0.237 & 0.624 &2.764&1.809\\
                              & Adj. Events &0.717&0.424&0.285& 0.293 & 0.139 & 0.432 &2.112&1.452\\
                              & Relations &0.642&\textbf{0.513}&\textbf{0.483}& \textbf{0.129} & \textbf{0.030} & \textbf{0.159} &1.220&0.415 \\ \hline
\multirow{3}{1.5cm}{Graph} & \rule{0pt}{1em} Full Story & 0.678 & 0.320 &0.127&0.358  & 0.193 & 0.551  & 2.783&2.070\\
                              & Adj. Story & \textbf{0.702} & \textbf{0.441} & \textbf{0.202} & \textbf{0.261} & 0.239 & \textbf{0.500} &2.143&1.567  \\
                              & Ext. Story & \textbf{0.702} & 0.356 &0.184 & 0.346 & \textbf{0.172} &0.518 & 2.573&1.886  \\ \hline
\end{tabular}
\caption{Effects of narrative representation and distance metric type on Krippendorff’s $\alpha$ for narrative extraction task. $\Delta$ denotes the difference in $\alpha$ between lenient, moderate and strict metrics. $|\cdot|$ indicates the average set cardinality (i.e., the number of events, relations, or triples per annotation). $\mu(\cdot)$ and $std(\cdot)$ denote mean and standard deviation operation.}
\label{tab2}
\end{table*}

As Task~2 is formulated as a graph-based annotation task, we employed a $6\times3$ factorial experimental design to examine the effects of narrative representation (six levels) and distance metric type (three levels) on narrative annotation reliability. Since only a single set of IAA measurements was available, we report only descriptive statistics, instead of providing significance tests\footnote{Typically, a $6\times3$ factorial design requires at least 15 samples per group to enable ANOVA.}.

The first factor, narrative representation (see Fig.~\ref{fig3}), spans six variants capturing different levels of abstraction and structural complexity. Three are categorical representations: 
(1) \textbf{All Events}, a set of node labels in a multi-hop subgraph reachable \textit{from and to} the target node \textit{Inflation}, including all events identified by annotators\footnote{A set of node labels in \textit{Full Story}.}; 
(2) \textbf{Adjacent Events}, a set of node labels in a one-hop subgraph reachable \textit{to Inflation}, limited to annotated events causing a change in \textit{Inflation}\footnote{A set of node labels in \textit{Adjacent Story}.}; and 
(3) \textbf{Relations}, a set of edge labels in Adjacent Story (described below), representing annotators' perceived direction of change in \textit{Inflation}\footnote{A set of relation labels in \textit{Adjacent Story}.}. 
The remaining three are graph-based representations:
(4) \textbf{Full Story}, a multi-hop subgraph encompassing all annotated events and relations in triple forms $\{\{e_i, r_{ij}, e_j\}, \{e_j, r_{jk}, e_k\}\}$; 
(5) \textbf{Adjacent Story}, a one-hop subgraph of \textit{Full Story}, restricted to events and their relations directly causing a change in \textit{Inflation}; and 
(6) \textbf{Extended Story}, a multi-hop subgraph extended from \textit{Adjacent Story}, including events and their relations indirectly causing a change in \textit{Inflation}.

The second factor, distance metric type, consists of three types: 
(1) \textbf{Lenient metric}, based on partial overlaps between annotations (Section~\ref{sec:lenient}); 
(2) \textbf{Moderate metric}, which captures graded similarity by considering the proportion of shared elements between annotations (Section~\ref{sec:moderate}), and 
(3) \textbf{Strict metric}, which requires exact matches in a set of nodes, edges or triples (Section~\ref{sec:strict}). 

It is important to note that IAA scores for nodes and edges are not directly comparable to those for graphs, since categorical features and graph features measure different aspects of narratives, and they require different distance metrics.
On the other hand, a comparison of IAA scores between categorical representations or between graph representations describes the influence of representation granularity to annotation variability.

A closer examination of Table~\ref{tab2} reveals three key observations regarding the effects of narrative representation and distance metric type on IAA.

First, \textbf{lenient metrics overestimate reliability in narrative annotation for both categorical and graph-based representation.}
For example, the notably high $\alpha$ for \textit{All Events} (0.861) with the lenient metric, followed by a sharp decline under the moderate metric (0.481, $\Delta_{ls}$=0.387) and strict metric (0.244, $\Delta_{ls}$=0.624), indicates that overlap-based lenient metrics tend to overestimate agreement for dense, high-coverage annotations, which increase incidental overlaps between annotators. As a result, lenient metric masks the difference in how annotators identify events and interpret causal relations. This decline of $\alpha$ is observed across all narrative representations, indicating stricter metrics reveals and penalizes more types of dissimilarity, e.g., mismatched or missing nodes and edges. Therefore, we encourage reporting multiple reliability scores across distance metrics of varying granularity in narrative graph annotation task, offering an informative overview of the dataset.

Second, \textbf{locally-constrained representations yields higher annotation consistency (or lower annotation variability).}
Annotators show higher agreement and consistency when the annotation scope is limited to local structures. The \textit{Relations} category yields the highest moderate (0.513) and strict (0.483) $\alpha$, while having the smallest declines ($\Delta_{lm}$=0.129, $\Delta_{ms}$=0.030, $\Delta_{ls}$=0.159), indicating that annotators more consistently identify directional relations than events or graphs. Similarly, the \textit{Adjacent Story} achieves the most balanced agreement among graph-based representations (lenient: 0.702, moderate: 0.441, strict: 0.202). In other words, its relatively high reliability score experiences a less-pronounced decline. Notably, higher annotation consistency is also reflected in lower standard deviation of annotation cardinality (i.e., the number of nodes, edges or triples per annotation) for \textit{Relations} (0.415) and \textit{Adjacent Story} (1.567). On the contrary, full annotation coverage introduces the largest mean cardinality and cardinality spread (\textit{All Events}: $\mu$=2.764, $std$=1.809; \textit{Full Story}: $\mu$=2.783, $std$=2.070), which exhibits the largest lenient to strict evaluation gaps ($\Delta_{ls}$). This highlights a trade-off between coverage of narrative graph annotation and inter-annotator consistency.

Third, \textbf{Adjacent Story provides the most reliable and consistent graph representation.} 
Among graph-based variants, \textit{Adjacent Story} achieves the highest agreement across all distance metrics (lenient: 0.702, moderate: 0.441, strict: 0.202) while maintaining smaller or comparable gaps in reliability scores (e.g., $\Delta_{lm}$=0.261, $\Delta_{ms}$=0.239) compared to \textit{Full Story} and \textit{Extended Story}. In short, constraining graphs to immediate neighbors of the target node yields the best overall balance between contextual coverage and reliability robustness in our annotations, suggesting that \textit{Adjacent Story} can be considered a reliable and consistent graph-based representation for evaluating narrative extraction datasets.

\subsection{Where Do Annotators Disagree?}

To investigate label-specific disagreement in Task~1, 
we conduct an analysis of variance (ANOVA) to test whether the proportion of documents with full 
annotator agreement differs across majority-voted labels. In other words, we examine whether annotators 
achieve higher consensus for certain labels than for others.
As shown in Fig.~\ref{app:fig4}, articles with the majority label \textit{non-inflation-related} 
exhibit a substantially higher rate of full agreement ($\mu$=$0.770$, $\textit{sdt}$=$0.422$) compared to those 
labeled \textit{inflation-related} ($\mu$=$0.205$, $\textit{sdt}$=$0.405$) and \textit{inflation-cause-dominant} 
($\mu$=$0.317$, $\textit{sdt}$=$0.468$). The ANOVA results indicate a statistically significant difference 
across labels ($df$=$2$, $F$=$91.701^{***}$, $p$<$0.001$, $\eta^2$=$0.274$). Post-hoc pairwise Student’s \textit{t}-tests 
further show significant differences between \textit{inflation-related} and \textit{non-inflation-related} 
($t$=$-13.348^{***}$, $p$<$0.001$), between \textit{inflation-related} and \textit{inflation-cause-dominant} 
($t$=$-2.035^{*}$, $p$<$0.05$), and between \textit{non-inflation-related} and \textit{inflation-cause-dominant} 
($t$=$8.290^{***}$, $p$<$0.001$). These results suggest that annotators have greater difficulty reaching agreement 
on \textit{inflation-related} and \textit{inflation-cause-dominant} instances than on \textit{non-inflation-related} 
ones.
For Task~2, we analyze disagreement at the triple level by identifying, for each article, 
the least-agreed triples—that is, those annotated least frequently across annotators. 
For each article, we first compute the union of all annotated triples and then determine which triples have 
the lowest annotation frequency. Across 104 articles (each annotated by four annotators), the five least-agreed 
triples are: \textit{(Supply (residual), Increases, Inflation)}, \textit{(Monetary Policy, Increases, Inflation)}, 
\textit{(Energy Prices, Increases, Inflation)}, \textit{(Monetary Policy, Decreases, Inflation)}, and 
\textit{(Wages, Increases, Inflation)}. These findings should be interpreted as preliminary. 
Future work should develop more systematic and fine-grained methods to quantify and characterize annotator 
disagreement.

\section{Related work}

\subsection{Integrating QCA in NLP}
The integration of QCA as a methodological framework for data annotation remains limited in NLP.
~\citet{yu2011compatibility} provided epistemological arguments for the compatibility of computational and qualitative research methods, and the importance of iterative refinement of category system. ~\citet{DBLP:conf/fat/BirhaneKCADB22} applied QCA at a meta-research level, providing an analysis of research culture and values in ML papers. Another line of work attempts to automate QCA by training models to perform the coding process on textual data~\citep{liew-etal-2014-optimizing, grandeit-etal-2020-using, fischer-biemann-2024-exploring,  DBLP:conf/iui/OverneySDR24}. While this approach leverages computational efficiency, it differs from a \textbf{human-centered} QCA workflow, where iterative interpretation and discussion guide the annotation process. To our knowledge, no prior work has applied QCA as a methodological framework for data annotation in machine learning. This may be attributed to the inherently time-intensive process of QCA, which contrasts with the fast-paced, benchmark-driven culture typical of machine learning research.

\subsection{Evaluation on Narrative Graph Annotations}
The most similar work to our annotation evaluation framework is 
in NLP concerns using suitable distance metrics that account for the complex phenomena of semantic and pragmatic annotation. \textit{Measuring Agreement on Set-valued Items (MASI)} metric extends Jaccard similarity by assigning weights to different type of edit operations, applied in coreference resolution~\citep{DBLP:conf/lrec/Passonneau04} and document summarization~\citep{DBLP:conf/lrec/Passonneau06}. Building on this foundation, we introduce a family of distance metrics (lenient, moderate, and strict) that capture different granularities of reliability to better capture the complex phenomenon of narrative annotation.
\textit{Labeled Attachment Score (LAS)} and \textit{Unlabeled Attachment Score (UAS)} are distance metrics used in dependency parsing~\citep{DBLP:conf/conll/BuchholzM06}. While LAS provides a stricter, more fine-grained measure by requiring both the correct head and dependency label, UAS is a more lenient metric that only considers structural correctness. We draw a parallel between LAS/UAS and our proposal of providing lenient vs. moderate vs. strict metrics for narrative graph evaluation. However, unlike LAS/UAS which are used to assess parser performance, our metrics are designed to evaluate annotation quality in terms of providing multiple IAA scores under different conditions, a critical step in understanding human variation in complex tasks like narrative understanding.

\section{Conclusion}

In this work, we presented a narrative graph dataset on inflation narratives. 
We introduced a QCA-based annotation methodology to reduce annotation errors by iteratively refining our 
category system and annotation guideline. An important goal of this QCA-based annotation methodology is to 
establish a common-ground understanding of the task. 
To evaluate the annotation under HLV, we employed a $6\times3$ factorial experimental design to examine 
the effects of narrative representations and distance metrics on annotation reliability. 
Our analysis demonstrates that both the choice of narrative representation and distance metric 
strongly influence IAA. Specifically, while lenient, overlap-based distance metrics overestimate reliability 
in narrative annotation, stricter metrics revealed commonly-perceived narrative elements among annotators. 
This reflects the coexistence of multiple plausible readings of the same text. 
Furthermore, constraining annotation to local structures, as in \textit{Relations} and \textit{Adjacent Story}, 
improves consistency in reliability while retaining essential narrative elements. 
Multi-hop subgraph representations, while richer in context, reduce strict agreement and highlight a 
trade-off between contextual completeness and reliability. 
These findings provide practical guidance for the design and evaluation of narrative graph annotations 
and underscore the value of comparing multiple reliability scores to capture the complexity of narrative 
interpretation.

\section*{Acknowledgments}
The authors acknowledge the financial support by the Hub of Computing and Data Science (HCDS) of University of Hamburg within the Cross-Disciplinary Lab programme, as well as by the Institute of Information Systems (IIS) of the Leuphana University Lüneburg.

\section*{Ethics Statement and Broader Impacts}
To our knowledge, this work does not concern any substantial ethical issue. No private or sensitive data is used or displayed. Annotators were trained and informed about the research purpose. Identity of annotators is not revealed. The dataset and evaluation methodology can benefit applications in media narrative analysis, economic forecasting, and social science research, where understanding narratives is crucial. At the same time, users should be mindful that even with structured annotations, narratives reflect human interpretation and may not fully capture all perspectives or nuances in news reporting.

\section*{Limitations}

\subsection{Data, Annotators and Annotation Methodology}

In terms of data, the dataset used in this study is both limited in size and constrained by licensing, which may hinder reproducibility and broader adoption. Its relatively small scale also reduces statistical power and limits the generalizability of findings across domains or narrative genres.

In terms of annotators, on the one hand, the recruitment criteria could be improved by including a more diverse range of sociodemographic backgrounds. On the other hand, the limited number of annotators restricts our ability to perform rigorous statistical testing. Future work should include a larger and more diverse participant pool to support more robust experimental designs and statistically sound comparisons.

By deductively introducing our initial category system, rather than developing it solely inductively from the data, we predefine interpretations of events, which may lead to a higher level of disagreement among our annotators. Additionally, establishing shared understanding among annotators required ongoing negotiation of theoretical and methodological perspectives. This process is context-dependent and time-intensive, increasing the difficulty to achieve standardization ~\cite{Stamann2019}. 

\subsection{Localization of Disagreements}

Although we conduct quantitative analyses to identify where disagreement occurs, our approach to localizing disagreement remains relatively coarse-grained. For Task~1, disagreement is measured at the document level through majority labels and full-agreement proportions, which does not capture more nuanced patterns such as partial agreement, systematic pairwise disagreement, or uncertainty concentrated around specific linguistic constructions. As a result, the analysis may obscure fine-grained sources of ambiguity.

For Task~2, disagreement is measured with annotation frequency of triples. While this provides an initial indication of which triples are less consistently identified, it does not distinguish between different types of disagreement (e.g., structural differences between graphs vs.\ semantic differences between nodes or edges). 

Future work should therefore adopt more systematic methods for measuring disagreement.

\subsection{Narrative Representation and Distance Metrics}
In terms of narrative representation, DAGs effectively capture the semantic structure of narratives, such as events and their causal relationships, but fall short in representing narrative discourse features. These include the manner in which a story is told, e.g., aspects of tone, emotion, and cultural framing.
Naturally, the discussion of what the distance metrics are measuring arises when we speak about feature representation. Current evaluations of graph annotations do not consider the semantic similarity of node labels. This means that concepts with similar meanings but different labels are treated as completely different. This could result in annotators’ actual agreement being underestimated. Moreover, weighted edit-distance as in ~\citet{DBLP:conf/lrec/Passonneau06} can further granulate the notion of similarity, in the context of narrative graph. Future work should further explore the theory-driven measure of narrative similarity.

\subsection{Representing Core Narrative Elements}
A key question in narrative annotation is which representations effectively capture core narrative elements. Drawing from literary structuralism, the smallest functional units in narratives are events that cause a change of state. As ~\citet{abbottCambridgeIntroductionNarrative2002} note, narratives consist of constituent events, without which the story would fundamentally change or would not make sense. They are crucial to the plot's development, driving the narrative forward. While our results suggest that graph-based representations, such as \textit{Adjacent Story}, reflect some level of shared understanding among annotators, they do not provide a direct measure of core narrative elements. Addressing this limitation requires a more precise operationalization of what constitutes a ``core'' event within a narrative. Recent work in NLP has begun to explore this direction: ~\citet{DBLP:conf/acl/AntoniakMSAP24} proposed methods to extract salient elements from Reddit stories, while ~\citet{huang-usbeck-2024-narration} highlighted the importance of extracting "core story" and distinguished constituent events from supplementary events, which served a decorative purpose in narratives. Future work should build on these insights to define and evaluate core narrative elements more rigorously.

\section{Bibliographical References}\label{sec:reference}

\bibliographystyle{lrec2026-natbib}
\bibliography{lrec2026-example}

@book{akerlofAnimalSpiritsHow2010,
	address = {Princeton, N.J Woodstock},
	title = {Animal spirits: how human psychology drives the economy, and why it matters for global capitalism},
	isbn = {978-0-691-14592-1 978-1-4008-3472-3 978-0-691-14233-3},
	shorttitle = {Animal spirits},
	language = {eng},
	publisher = {Princeton University Press},
	author = {Akerlof, George A. and Shiller, Robert J.},
	year = {2010},
}

@techreport{Weinig2025Going,
	address = {Hamburg},
	type = {{WiSo}-{HH} {Working} {Paper} {Series}},
	title = {Going viral: {Inflation} narratives and the macroeconomy},
	url = {https://hdl.handle.net/10419/307613.2},
	language = {eng},
	number = {86},
	institution = {Universität Hamburg, Fakultät für Wirtschafts- und Sozialwissenschaften, WiSo-Forschungslabor},
	author = {Weinig, Max and Fritsche, Ulrich},
	year = {2025},
}

@inproceedings{parfenova-etal-2025-text,
    title = "Text Annotation via Inductive Coding: Comparing Human Experts to {LLM}s in Qualitative Data Analysis",
    author = {Parfenova, Angelina  and
      Marfurt, Andreas  and
      Pfeffer, J{\"u}rgen  and
      Denzler, Alexander},
    editor = "Chiruzzo, Luis  and
      Ritter, Alan  and
      Wang, Lu",
    booktitle = "Findings of the Association for Computational Linguistics: NAACL 2025",
    month = apr,
    year = "2025",
    address = "Albuquerque, New Mexico",
    publisher = "Association for Computational Linguistics",
    url = "https://aclanthology.org/2025.findings-naacl.361/",
    doi = "10.18653/v1/2025.findings-naacl.361",
    pages = "6471--6484",
    ISBN = "979-8-89176-195-7"
}

@inproceedings{heddaya-etal-2024-causal,
    title = "Causal Micro-Narratives",
    author = "Heddaya, Mourad  and
      Zeng, Qingcheng  and
      Zentefis, Alexander  and
      Voigt, Rob  and
      Tan, Chenhao",
    editor = "Lal, Yash Kumar  and
      Clark, Elizabeth  and
      Iyyer, Mohit  and
      Chaturvedi, Snigdha  and
      Brei, Anneliese  and
      Brahman, Faeze  and
      Chandu, Khyathi Raghavi",
    booktitle = "Proceedings of the 6th Workshop on Narrative Understanding",
    month = nov,
    year = "2024",
    address = "Miami, Florida, USA",
    publisher = "Association for Computational Linguistics",
    url = "https://aclanthology.org/2024.wnu-1.12/",
    doi = "10.18653/v1/2024.wnu-1.12",
    pages = "67--84"
}

@book{tuckettMindingMarkets2011,
  title = {Minding the {{Markets}}: An Emotional Finance View of Financial Instability},
  author = {Tuckett, David},
  year = {2011},
  publisher = {Palgrave Macmillan UK},
  location = {London},
  doi = {10.1057/9780230307827},
  url = {http://link.springer.com/10.1057/9780230307827},
  urldate = {2025-09-29},
  isbn = {978-1-349-33551-0 978-0-230-30782-7},
  langid = {english}
}

@article{ Stamann2019,
 title = {Becoming Able to Work Together as a Central Challenge for a Successful Joint Practice—Experiences From a Qualitative Content Analysis Interpretation Group and Suggestions for Arranging Interpretation Group Sessions },
 author = {Stamann, Christoph and Janssen, Markus},
 journal = {Forum Qualitative Sozialforschung: Qualitative Social Research},
 number = {3},
 volume = {20},
 year = {2019},
 issn = {1438-5627},
 doi = {https://doi.org/10.17169/fqs-20.3.3379},
 urn = {https://nbn-resolving.org/urn:nbn:de:0168-ssoar-65473-7},
}

@book{Mayring2014,
 title = {Qualitative content analysis: theoretical foundation, basic procedures and software solution},
 author = {Mayring, Philipp},
 year = {2014},
 pages = {143},
 address = {Klagenfurt},
 urn = {https://nbn-resolving.org/urn:nbn:de:0168-ssoar-395173}}

@article{gehringAnalyzingClimateChange2023,
	title = {Analyzing {Climate} {Change} {Policy} {Narratives} with the {Character}-{Role} {Narrative} {Framework}},
	issn = {1556-5068},
	url = {https://www.ssrn.com/abstract=4456361},
	doi = {10.2139/ssrn.4456361},
	language = {en},
	urldate = {2024-02-14},
	journal = {SSRN Electronic Journal},
	author = {Gehring, Kai and Grigoletto, Matteo},
	year = {2023},
}

@inproceedings{gueta-etal-2025-llms,
    title = "Can {LLM}s Learn Macroeconomic Narratives from Social Media?",
    author = "Gueta, Almog  and
      Feder, Amir  and
      Gekhman, Zorik  and
      Goldstein, Ariel  and
      Reichart, Roi",
    editor = "Chiruzzo, Luis  and
      Ritter, Alan  and
      Wang, Lu",
    booktitle = "Findings of the Association for Computational Linguistics: NAACL 2025",
    month = apr,
    year = "2025",
    address = "Albuquerque, New Mexico",
    publisher = "Association for Computational Linguistics",
    url = "https://aclanthology.org/2025.findings-naacl.4/",
    doi = "10.18653/v1/2025.findings-naacl.4",
    pages = "57--78",
    ISBN = "979-8-89176-195-7"
}

@article{ashRelatioTextSemantics2024,
	title = {Relatio: {Text} {Semantics} {Capture} {Political} and {Economic} {Narratives}},
	volume = {32},
	copyright = {https://creativecommons.org/licenses/by/4.0},
	issn = {1047-1987, 1476-4989},
	url = {https://www.cambridge.org/core/product/identifier/S1047198723000086/type/journal_article},
	doi = {10.1017/pan.2023.8},
	language = {en},
	number = {1},
	urldate = {2025-03-28},
	journal = {Political Analysis},
	author = {Ash, Elliott and Gauthier, Germain and Widmer, Philine},
	month = jan,
	year = {2024},
	pages = {115--132},
}

@article{terellenNarrativeMonetaryPolicy2022,
	title = {Narrative {Monetary} {Policy} {Surprises} and the {Media}},
	volume = {54},
	issn = {0022-2879, 1538-4616},
	url = {https://onlinelibrary.wiley.com/doi/10.1111/jmcb.12868},
	doi = {10.1111/jmcb.12868},
	language = {en},
	number = {5},
	urldate = {2024-09-18},
	journal = {Journal of Money, Credit and Banking},
	author = {Ter Ellen, Saskia and Larsen, Vegard H. and Thorsrud, Leif Anders},
	month = aug,
	year = {2022},
	pages = {1525--1549},
}

@article{müllerGermanInflationNarrative2022,
	title = {A {German} {Inflation} {Narrative}},
	url = {https://eldorado.tu-dortmund.de/handle/2003/40775},
	doi = {10.17877/DE290R-22632},
	language = {en},
	urldate = {2024-02-13},
	journal = {DoCMA Working Paper;9},
	author = {Müller, Henrik and Schmidt, Tobias and Rieger, Jonas and Hufnagel, Lena Maria and Hornig, Nico},
	collaborator = {{Technische Universität Dortmund} and {Technische Universität Dortmund}},
	month = mar,
	year = {2022},
	note = {Publisher: TU Dortmund},
}

@incollection{selviQualitativeContentAnalysis2019,
	address = {New York : Taylor and Francis, 2020. Series: Routledge handbooks in applied linguistics},
	edition = {1},
	title = {Qualitative content analysis},
	isbn = {978-0-367-82447-1},
	url = {https://www.taylorfrancis.com/books/9781000734034},
	language = {en},
	urldate = {2025-05-19},
	booktitle = {The {Routledge} {Handbook} of {Research} {Methods} in {Applied} {Linguistics}},
	publisher = {Routledge},
	author = {Selvi, Ali Fuad},
	editor = {McKinley, Jim and Rose, Heath},
	month = dec,
	year = {2019},
	doi = {10.4324/9780367824471},
}

@book{beckertUncertainFuturesImaginaries2018,
	address = {Oxford New York, NY},
	edition = {First edition},
	title = {Uncertain futures: imaginaries, narratives, and calculation in the economy},
	isbn = {978-0-19-882080-2},
	shorttitle = {Uncertain futures},
	language = {eng},
	publisher = {Oxford University Press},
	editor = {Beckert, Jens and Bronk, Richard},
	year = {2018},
}

@article{pollettaSociologyStorytelling2011,
	title = {The {Sociology} of {Storytelling}},
	volume = {37},
	issn = {0360-0572, 1545-2115},
	url = {https://www.annualreviews.org/doi/10.1146/annurev-soc-081309-150106},
	doi = {10.1146/annurev-soc-081309-150106},
	language = {en},
	number = {1},
	urldate = {2024-10-11},
	journal = {Annual Review of Sociology},
	author = {Polletta, Francesca and Chen, Pang Ching Bobby and Gardner, Beth Gharrity and Motes, Alice},
	month = aug,
	year = {2011},
	pages = {109--130},
}

@book{abbottCambridgeIntroductionNarrative2002,
	address = {Cambridge, UK; New York, NY, USA},
	title = {The {Cambridge} introduction to narrative},
	isbn = {978-0-521-65033-5 978-0-521-65969-7},
	language = {en},
	publisher = {Cambridge University Press},
	author = {Abbott, H. Porter},
	year = {2002},
}

@article{shillerNarrativeEconomics2017,
	title = {Narrative {Economics}},
	volume = {107},
	issn = {0002-8282},
	url = {https://pubs.aeaweb.org/doi/10.1257/aer.107.4.967},
	doi = {10.1257/aer.107.4.967},
	language = {en},
	number = {4},
	urldate = {2023-07-22},
	journal = {American Economic Review},
	author = {Shiller, Robert J.},
	month = apr,
	year = {2017},
	pages = {967--1004},
}

@article{roosNarrativesEconomics2023,
	title = {Narratives in economics},
	issn = {0950-0804, 1467-6419},
	url = {https://onlinelibrary.wiley.com/doi/10.1111/joes.12576},
	doi = {10.1111/joes.12576},
	language = {en},
	urldate = {2024-02-02},
	journal = {Journal of Economic Surveys},
	author = {Roos, Michael and Reccius, Matthias},
	month = jun,
	year = {2023},
	pages = {joes.12576},
}

@article{10.1093/restud/rdag014,
    author = {Andre, Peter and Haaland, Ingar and Roth, Christopher and Wiederholt, Mirko and Wohlfart, Johannes},
    title = {Narratives about the Macroeconomy*},
    journal = {The Review of Economic Studies},
    pages = {rdag014},
    year = {2026},
    month = {02},
    issn = {0034-6527},
    doi = {10.1093/restud/rdag014},
    url = {https://doi.org/10.1093/restud/rdag014},
    eprint = {https://academic.oup.com/restud/advance-article-pdf/doi/10.1093/restud/rdag014/66967771/rdag014.pdf},
}

@Book{Pearl.2009,
  author    = {Pearl, Judea},
  publisher = {{Cambridge University Press}},
  title     = {Causality: Models, reasoning, and inference},
  year      = {2009},
  address   = {Cambridge},
  edition   = {2},
}

@article{eliazModelCompetingNarratives2020,
	title = {A {Model} of {Competing} {Narratives}},
	volume = {110},
	issn = {0002-8282},
	url = {https://pubs.aeaweb.org/doi/10.1257/aer.20191099},
	doi = {10.1257/aer.20191099},
	language = {en},
	number = {12},
	urldate = {2023-07-22},
	journal = {American Economic Review},
	author = {Eliaz, Kfir and Spiegler, Ran},
	month = dec,
	year = {2020},
	pages = {3786--3816},
}

@incollection{przyborksiForschungsdesignFuerQualitative2022,
	address = {Wiesbaden},
	title = {Forschungsdesign für die qualitative {Sozialforschung}},
	copyright = {https://www.springer.com/tdm},
	isbn = {978-3-658-37984-1 978-3-658-37985-8},
	url = {https://link.springer.com/10.1007/978-3-658-37985-8},
	language = {de},
	urldate = {2025-05-15},
	booktitle = {Handbuch {Methoden} der empirischen {Sozialforschung}},
	publisher = {Springer Fachmedien Wiesbaden},
	author = {Przyborksi, Aglaja and Wohlrab-Sahr},
	editor = {Baur, Nina and Blasius, Jörg},
	year = {2022},
	doi = {10.1007/978-3-658-37985-8},
}

@inproceedings{huang-usbeck-2024-narration,
    title = "Narration as Functions: from Events to Narratives",
    author = "Huang, Junbo  and
      Usbeck, Ricardo",
    editor = "Lal, Yash Kumar  and
      Clark, Elizabeth  and
      Iyyer, Mohit  and
      Chaturvedi, Snigdha  and
      Brei, Anneliese  and
      Brahman, Faeze  and
      Chandu, Khyathi Raghavi",
    booktitle = "Proceedings of the 6th Workshop on Narrative Understanding",
    month = nov,
    year = "2024",
    address = "Miami, Florida, USA",
    publisher = "Association for Computational Linguistics",
    url = "https://aclanthology.org/2024.wnu-1.1/",
    doi = "10.18653/v1/2024.wnu-1.1",
    pages = "1--7"
}

@inproceedings{DBLP:conf/acl/AntoniakMSAP24,
  author       = {Maria Antoniak and
                  Joel Mire and
                  Maarten Sap and
                  Elliott Ash and
                  Andrew Piper},
  editor       = {Lun{-}Wei Ku and
                  Andre Martins and
                  Vivek Srikumar},
  title        = {Where Do People Tell Stories Online? Story Detection Across Online
                  Communities},
  booktitle    = {Proceedings of the 62nd Annual Meeting of the Association for Computational
                  Linguistics (Volume 1: Long Papers), {ACL} 2024, Bangkok, Thailand,
                  August 11-16, 2024},
  pages        = {7104--7130},
  publisher    = {Association for Computational Linguistics},
  year         = {2024},
  url          = {https://doi.org/10.18653/v1/2024.acl-long.383},
  doi          = {10.18653/V1/2024.ACL-LONG.383},
  bibsource    = {dblp computer science bibliography, https://dblp.org}
}

@misc{krippendorff2011computing,
  title={Computing Krippendorff’s alpha-reliability},
  journal={Departmental Papers (ASC)},
  author={Krippendorff, Klaus},
  year={2011},
  url={https://repository.upenn.edu/handle/20.500.14332/2089}
}

@inproceedings{DBLP:conf/naacl/ZaratianaTHC24,
  author       = {Urchade Zaratiana and
                  Nadi Tomeh and
                  Pierre Holat and
                  Thierry Charnois},
  editor       = {Kevin Duh and
                  Helena G{\'{o}}mez{-}Adorno and
                  Steven Bethard},
  title        = {GLiNER: Generalist Model for Named Entity Recognition using Bidirectional
                  Transformer},
  booktitle    = {Proceedings of the 2024 Conference of the North American Chapter of
                  the Association for Computational Linguistics: Human Language Technologies
                  (Volume 1: Long Papers), {NAACL} 2024, Mexico City, Mexico, June 16-21,
                  2024},
  pages        = {5364--5376},
  publisher    = {Association for Computational Linguistics},
  year         = {2024},
  url          = {https://doi.org/10.18653/v1/2024.naacl-long.300},
  doi          = {10.18653/V1/2024.NAACL-LONG.300},
  bibsource    = {dblp computer science bibliography, https://dblp.org}
}

@inproceedings{DBLP:conf/emnlp/Plank22,
  author       = {Barbara Plank},
  editor       = {Yoav Goldberg and
                  Zornitsa Kozareva and
                  Yue Zhang},
  title        = {The "Problem" of Human Label Variation: On Ground Truth in Data, Modeling
                  and Evaluation},
  booktitle    = {Proceedings of the 2022 Conference on Empirical Methods in Natural
                  Language Processing, {EMNLP} 2022, Abu Dhabi, United Arab Emirates,
                  December 7-11, 2022},
  pages        = {10671--10682},
  publisher    = {Association for Computational Linguistics},
  year         = {2022},
  url          = {https://doi.org/10.18653/v1/2022.emnlp-main.731},
  doi          = {10.18653/V1/2022.EMNLP-MAIN.731},
  bibsource    = {dblp computer science bibliography, https://dblp.org}
}

@inproceedings{DBLP:conf/fat/BirhaneKCADB22,
  author       = {Abeba Birhane and
                  Pratyusha Kalluri and
                  Dallas Card and
                  William Agnew and
                  Ravit Dotan and
                  Michelle Bao},
  title        = {The Values Encoded in Machine Learning Research},
  booktitle    = {FAccT '22: 2022 {ACM} Conference on Fairness, Accountability, and
                  Transparency, Seoul, Republic of Korea, June 21 - 24, 2022},
  pages        = {173--184},
  publisher    = {{ACM}},
  year         = {2022},
  url          = {https://doi.org/10.1145/3531146.3533083},
  doi          = {10.1145/3531146.3533083},
  bibsource    = {dblp computer science bibliography, https://dblp.org}
}

@article{yu2011compatibility,
  title={Compatibility between text mining and qualitative research in the perspectives of grounded theory, content analysis, and reliability.},
  author={Yu, Chong Ho and Jannasch-Pennell, Angel and DiGangi, Samuel},
  journal={Qualitative Report},
  volume={16},
  number={3},
  pages={730--744},
  year={2011},
  publisher={ERIC},
  doi = {https://doi.org/10.46743/2160-3715/2011.1085},
}

@inproceedings{grandeit-etal-2020-using,
    title = "Using {BERT} for Qualitative Content Analysis in Psychosocial Online Counseling",
    author = "Grandeit, Philipp  and
      Haberkern, Carolyn  and
      Lang, Maximiliane  and
      Albrecht, Jens  and
      Lehmann, Robert",
    editor = "Bamman, David  and
      Hovy, Dirk  and
      Jurgens, David  and
      O'Connor, Brendan  and
      Volkova, Svitlana",
    booktitle = "Proceedings of the Fourth Workshop on Natural Language Processing and Computational Social Science",
    month = nov,
    year = "2020",
    address = "Online",
    publisher = "Association for Computational Linguistics",
    url = "https://aclanthology.org/2020.nlpcss-1.2/",
    doi = "10.18653/v1/2020.nlpcss-1.2",
    pages = "11--23"
}

@inproceedings{fischer-biemann-2024-exploring,
    title = "Exploring Large Language Models for Qualitative Data Analysis",
    author = "Fischer, Tim  and
      Biemann, Chris",
    editor = {H{\"a}m{\"a}l{\"a}inen, Mika  and
      {\"O}hman, Emily  and
      Miyagawa, So  and
      Alnajjar, Khalid  and
      Bizzoni, Yuri},
    booktitle = "Proceedings of the 4th International Conference on Natural Language Processing for Digital Humanities",
    month = nov,
    year = "2024",
    address = "Miami, USA",
    publisher = "Association for Computational Linguistics",
    url = "https://aclanthology.org/2024.nlp4dh-1.41/",
    doi = "10.18653/v1/2024.nlp4dh-1.41",
    pages = "423--437"
}

@inproceedings{liew-etal-2014-optimizing,
    title = "Optimizing Features in Active Machine Learning for Complex Qualitative Content Analysis",
    author = "Liew, Jasy Suet Yan  and
      McCracken, Nancy  and
      Zhou, Shichun  and
      Crowston, Kevin",
    editor = "Danescu-Niculescu-Mizil, Cristian  and
      Eisenstein, Jacob  and
      McKeown, Kathleen  and
      Smith, Noah A.",
    booktitle = "Proceedings of the {ACL} 2014 Workshop on Language Technologies and Computational Social Science",
    month = jun,
    year = "2014",
    address = "Baltimore, MD, USA",
    publisher = "Association for Computational Linguistics",
    url = "https://aclanthology.org/W14-2513/",
    doi = "10.3115/v1/W14-2513",
    pages = "44--48"
}

@inproceedings{DBLP:conf/iui/OverneySDR24,
  author       = {Cassandra Overney and
                  Bel{\'{e}}n Sald{\'{\i}}as and
                  Dimitra Dimitrakopoulou and
                  Deb Roy},
  title        = {SenseMate: An Accessible and Beginner-Friendly Human-AI Platform for
                  Qualitative Data Analysis},
  booktitle    = {Proceedings of the 29th International Conference on Intelligent User
                  Interfaces, {IUI} 2024, Greenville, SC, USA, March 18-21, 2024},
  pages        = {922--939},
  publisher    = {{ACM}},
  year         = {2024},
  url          = {https://doi.org/10.1145/3640543.3645194},
  doi          = {10.1145/3640543.3645194},
  bibsource    = {dblp computer science bibliography, https://dblp.org}
}

@inproceedings{DBLP:conf/conll/BuchholzM06,
  author       = {Sabine Buchholz and
                  Erwin Marsi},
  editor       = {Llu{\'{\i}}s M{\`{a}}rquez and
                  Dan Klein},
  title        = {CoNLL-X Shared Task on Multilingual Dependency Parsing},
  booktitle    = {Proceedings of the Tenth Conference on Computational Natural Language
                  Learning, CoNLL 2006, New York City, USA, June 8-9, 2006},
  pages        = {149--164},
  publisher    = {{ACL}},
  year         = {2006},
  url          = {https://aclanthology.org/W06-2920/},
  bibsource    = {dblp computer science bibliography, https://dblp.org}
}

@inproceedings{DBLP:conf/lrec/Passonneau06,
  author       = {Rebecca J. Passonneau},
  editor       = {Nicoletta Calzolari and
                  Khalid Choukri and
                  Aldo Gangemi and
                  Bente Maegaard and
                  Joseph Mariani and
                  Jan Odijk and
                  Daniel Tapias},
  title        = {Measuring Agreement on Set-valued Items {(MASI)} for Semantic and
                  Pragmatic Annotation},
  booktitle    = {Proceedings of the Fifth International Conference on Language Resources
                  and Evaluation, {LREC} 2006, Genoa, Italy, May 22-28, 2006},
  pages        = {831--836},
  publisher    = {European Language Resources Association {(ELRA)}},
  year         = {2006},
  url          = {http://www.lrec-conf.org/proceedings/lrec2006/summaries/636.html},
  bibsource    = {dblp computer science bibliography, https://dblp.org}
}

@inproceedings{DBLP:conf/lrec/Passonneau04,
  author       = {Rebecca J. Passonneau},
  title        = {Computing Reliability for Coreference Annotation},
  booktitle    = {Proceedings of the Fourth International Conference on Language Resources
                  and Evaluation, {LREC} 2004, May 26-28, 2004, Lisbon, Portugal},
  publisher    = {European Language Resources Association},
  year         = {2004},
  url          = {http://www.lrec-conf.org/proceedings/lrec2004/summaries/752.htm},
  bibsource    = {dblp computer science bibliography, https://dblp.org}
}

\section{Appendices}
\subsection{Further Annotator Recruitment Information}
\label{app:annotators}
The students involved in the pilot phase are student assistants already affiliated with our research group. They are compensated according to the standard student assistant payment at the time of the study: €13.52 per hour (without a bachelor’s degree) and €15.21 per hour (with a bachelor’s degree or higher). 

For the final study, we employ four additional students under short-term service contracts. To ensure fair compensation and promote high-quality annotation, we provide payment above standard student rates. Each annotator receives a total of €912.60. Given 488 documents annotated in Task~1 (expected time: 4 minutes per document) and 104 documents in Task~2 (expected time: 10 minutes per document), the total estimated workload amounts to approximately 50 hours. This corresponds to an effective hourly wage of €18.30, exceeding both the standard student assistant wage at our institution and the country-level minimum wage by over €6 per hour. 

\subsection{Descriptive Statistics of the Annotated Dataset}
\label{app:descriptives}
\begin{figure}
\centering
\includegraphics[scale=0.45]{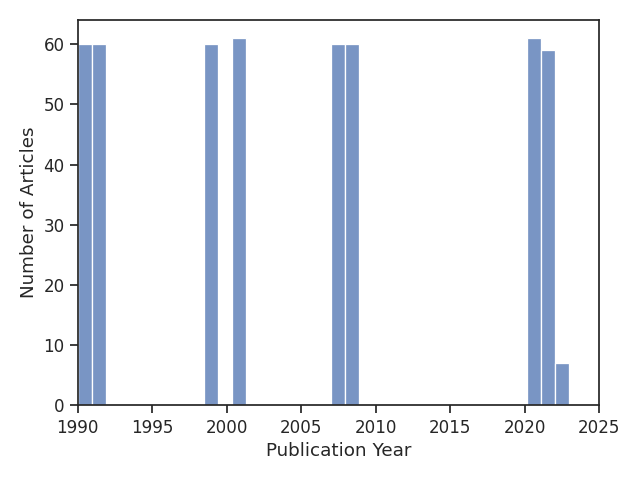}
\caption{Distribution of documents across publication years. 
The sampling strategy focuses on inflation-peak years: 1990, 1991, 1999, 2001, 2007, 2008, 2021, 2022 and 2023.}
\label{app:fig1}
\end{figure}

\begin{figure}
\centering
\includegraphics[scale=0.5]{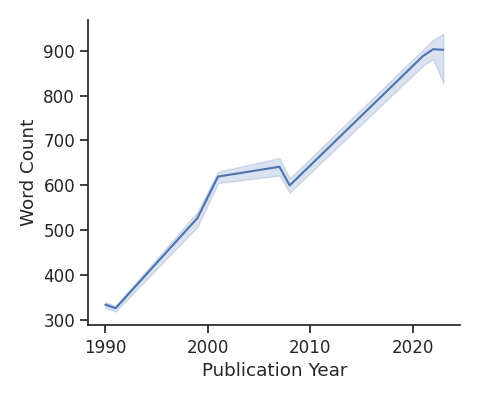}
\caption{Average word count of documents across publication years.}
\label{app:fig2}
\end{figure}

\begin{figure}
\centering
\includegraphics[scale=0.3]{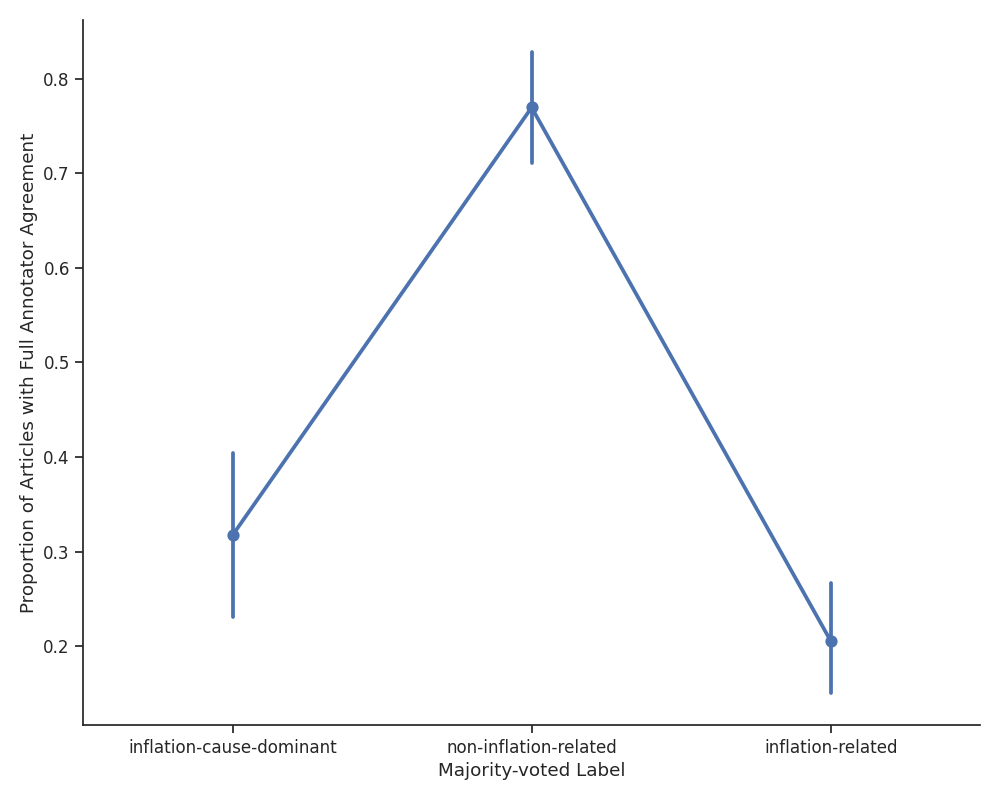}
\caption{Proportion of Articles with Full Annotators Agreement per Label.}
\label{app:fig4}
\end{figure}

The dataset consists of 488 documents, with an average length of 609 words (min: 112, max: 1774, median: 541). 
Articles are attributed with their publication year, as shown in Fig.~\ref{app:fig1}. The sampling strategy focuses 
on inflation-peak years: 1990, 1991, 1999, 2001, 2007, 2008, 2021, 2022 and 2023 (only until January). Fig.~\ref{app:fig2} 
shows an increase of word counts through time. 

\subsection{Annotation Example}
\label{app:example}
Snippet of a sample article:
\begin{quote}
ADB Exec Calls For Tighter Monetary Policies In Asia

  KUALA LUMPUR (AP)--Asia's growth is being threatened by spiraling inflation due to higher food and fuel cost, an Asian Development Bank executive warned Sunday, as he called on governments to tighten monetary policies to deal with the scourge. 

  The bank is reviewing its growth forecast of 7.6\% this year for Asia, excluding Japan, amid concerns inflation will widen income inequality in the region and cause more people to plunge into poverty, said its managing director general Rajat M.Nag. 

  \dots
\end{quote}

\begin{quote}
Annotator 1: 

\textit{[('Energy Prices', 'Increases', 'Inflation'), ('Monetary Policy', 'Decreases', 'Inflation'), ('Food Prices', 'Increases', 'Inflation')]}

Annotator 2: 

\textit{[('Energy Prices', 'Increases', 'Inflation'), ('Monetary Policy', 'Decreases', 'Inflation'), ('Food Prices', 'Increases', 'Inflation')]}

Annotator 3: 

\textit{[('Energy Prices', 'Increases', 'Inflation'), ('Food Prices', 'Increases', 'Inflation')]}

Annotator 4: 

\textit{[('Energy Prices', 'Increases', 'Inflation'), ('Government Spending', 'Increases', 'Inflation'), ('Food Prices', 'Increases', 'Inflation')]}
\end{quote}

\subsection{Annotation Guidelines}

Objective: Identify inflation narratives in large text corpora using a two-stage annotation process. \vspace{0.2cm}

\noindent \textbf{Stage I: Document-level Classification}\\ 
Determine whether the article addresses the cause of inflation. There are three cases: 

\begin{itemize}
        \item Inflation-cause-dominant: The text discusses the causes of inflation.
        \item Inflation-related: The text mentions inflation but not inflation causes. 
        \item Non-inflation-related: The text neither mentions causes of inflation nor inflation.
    \end{itemize}

\noindent \textbf{What is inflation?}
The rate of inflation indicates how much prices in a country rise from year to year, measured in percent. It is defined as the annual growth in the general price level of goods and services.
 A decline in the price level is commonly referred to as deflation. For example, an inflation rate of 2\% means that consumer prices rise by 2\% over 12 months. A basket of typical monthly purchases that costs €1,000 at the beginning of the year costs €1,020 at the end of the year.
Therefore, texts that mention solely movements in single prices (e.g. crude oil or wheat prices) are not to be labeled as “inflation-related” or “inflation-cause-dominant”. Results from this stage are reviewed to select documents for further annotation.\\ 

\noindent \textbf{Stage II: Text-Span Annotation} 

\noindent For documents labeled as Inflation-cause-dominant, identify text spans that indicate event mentions related to inflation causes. An event mention is a specific instance in text that describes an action, occurrence, or state change. It typically includes key information like the action itself (often expressed by a verb and/or noun), involved participants, and sometimes a time or place. Each event mentioned reflects one occurrence of an event as described in the text.
A selection of possible events are shown in Table 1. We distinguish between Demand, Supply, and Miscellaneous Events. \\
When highlighting the event spans:
\begin{itemize}
    \item Highlight only the minimum span (single/few token if possible)
    \item Usually, events include verbs and nouns, and sometimes adjectives
    \item If you are uncertain, always make a note and discuss it in the plenary sessions
\end{itemize}

To support event detection and increase the number of matching text spans, we pre-annotated the text using a pre-trained Named Entity Recognition (NER) zero-shot language model that outperforms general-purpose models like ChatGPT. However, due to potential prediction errors, human validation is essential . Specifically, some events might not be identified correctly or may not be labeled at all.

Furthermore, we aim to identify the narrated causal structure of the narratives. Therefore, you need to establish the causal relationships between the annotated events using Label Studio. To distinguish the effect direction, we include two relational attributes: Decreases and Increases.\\

\subsection{Category System for QCA}
\label{appendix:categories}
\begin{table*}[!ht]
	\centering
	\begin{tabular}{c|r|p{8cm}}		
    \hline
	\rule{0pt}{1em}\textbf{Category} & \textbf{Subcategory} &\textbf{Explanation} \\ \hline
	\multirow{5}{*}{\textbf{Demand}} & \rule{0pt}{1em} Government Spending  & Adjustments in government spending (e.g., stimulus payments) \\ 
			& Monetary Policy & Monetary policy by the Federal Reserve or other Central Banks \\ 
            &Pent-up Demand & Reopening of the economy and the associated higher incomes, new spending opportunities, and optimism about the future  \\ 
            &Demand Shift & Shift of demand across sectors (particularly increases in durables)  \\ 
			&Demand (Residuals) & Increase in demand that cannot be attributed to the other channels   \\ \hline
    \multirow{8}{*}{\textbf{Supply}}&  \rule{0pt}{1em}Supply Chain Issues   & Disruption of global supply chains \\ 
            &Transportation Costs & Rising shipping and freight costs, including container shortages and port congestion \\
			&Labor Shortage           & Shortage of workers, e.g., due to some workers dropping out of the labor force, and higher wage costs \\
            &Wages & Changes in wage levels driven by labor market dynamics, such as increased labor demand or worker bargaining power \\
			&Energy Prices & Higher energy prices, e.g., due to the global energy crisis, leading to shortages of oil and natural gas \\
            &Food Prices & Increases in food prices, e.g., driven by rising input costs, or global agricultural disruptions \\ 
            &Housing Costs & Rising housing and rental costs\\
			&Supply (Residual) & Other negative supply effects \\ \hline
	\multirow{13}{*}{\textbf{Miscellaneous}}   & \rule{0pt}{1em}Pandemic  & The COVID-19 pandemic, the global pandemic recession, lockdowns, and other policy measures \\ 
			&Politics & Policy failure, and mismanagement by policymakers, policymakers are blamed \\ 
			&War & Armed conflict involving state or non-state actors, such as the Russian invasion of Ukraine and the associated international economic, political, and military responses \\
            &Inflation Expectations & Expectations about high inflation in the coming years, making firms preemptively increase prices and workers bargain for higher wage \\
            &Base Effect & Mentions that inflation is high due to a base effect, i.e., a very low inflation rate during the pandemic, leading almost mechanically to high inflation rates now \\ 
			&Government Debt & High level of government debt \\
			&Tax Increases   & Tax increases, such as VAT hikes \\ 
            &Trade Balance & Inflationary pressures linked to changes in exports, imports, or trade deficits \\ 
            &Exchange Rates & Effects of exchange rate fluctuations on import prices and domestic inflation \\ 
            &Medical Costs & Rising healthcare or insurance costs, including structural issues in medical pricing \\ 
            &Education Costs & Increases in education-related expenses such as tuition fees\\ 
            &Climate Crisis & All aspects related to the climate crisis and natural disasters, as well as related environmental and economic consequences \\ 
			&Price-Gouging  & Companies exploit opportunities to increase profits \\ \hline                                                               
		\end{tabular}
        \caption{Events and Explanations in the Category System.}
\end{table*}
\end{document}